%% file: 0_main.tex
\documentclass[runningheads]{llncs}
\usepackage[T1]{fontenc}
\usepackage{graphicx}

\usepackage{amssymb}%add_yamaoka for eq
\usepackage{amsmath}%add_yamaoka for eq
\usepackage[whole]{bxcjkjatype}%For Japanese
\usepackage{booktabs}%For Table
\input{99_MatsushitaLabTemplate}%package of M lab

\begin{document}
%
%======================title and author======================%
%
\title{Learning from Similarity Proportion Loss for Classifying Skeletal Muscle Recovery Stages}
\author{Yu Yamaoka\inst{1}, 
Weng Ian Chan\inst{1}, 
Shigeto Seno\inst{1},\\
Soichiro Fukada\inst{2}, 
Hideo Matsuda\inst{1}
} %index{Last Name, First Name}
\authorrunning{Yu Yamaoka, Weng Ian Chan et al. }
\titlerunning{Learning from Similarity Proportion Loss}

\renewcommand{\thefootnote}{\fnsymbol{footnote}}
\institute{
    Graduate School of Information Science and Technology, \\ Osaka University, Osaka, Japan\\
    \email{\{yu-yamaoka, chan.wengian, senoo\}@ist.osaka-u.ac.jp} \and
    Graduate School of Pharmaceutical Sciences, Osaka University, Osaka, Japan
    \footnote[0]{Yu Yamaoka and Weng Ian Chan contributed equally to this work.}
}

%\orcidID{0009-0005-2527-1450}, \orcidID{0000-0003-3444-9431}
%Osaka University Information Science and Technology, Yamadaoka 1-5, Suita city, Osaka, Japan

\maketitle              % typeset the header of the contribution
%
%======================abstract======================%
%150～250 word
\begin{abstract}
Evaluating the regeneration process of damaged muscle tissue is a fundamental analysis in muscle research to measure experimental effect sizes and uncover mechanisms behind muscle weakness due to aging and disease. The conventional approach to assessing muscle tissue regeneration involves whole-slide imaging and expert visual inspection of the recovery stages based on the morphological information of cells and fibers. There is a need to replace these tasks with automated methods incorporating machine learning techniques to ensure a quantitative and objective analysis.
%To ensure quantitative and objective analysis, these tasks should be replaced by automated methods including machine learning. 
Given the limited availability of fully labeled data, a possible approach is Learning from Label Proportions (LLP), a weakly supervised learning method using class label proportions. However, current LLP methods have two limitations: (1) they cannot adapt the feature extractor for muscle tissues, and (2) they treat the classes representing recovery stages and cell morphological changes as nominal, resulting in the loss of ordinal information.
%the class, which represents the stages of recovery and the morphological changes of cells over time, as a nominal scale, resulting in the loss of ordinal information. 
To address these issues, we propose Ordinal Scale Learning from Similarity Proportion (OSLSP), which uses a similarity proportion loss derived from two bag combinations. OSLSP can update the feature extractor by using class proportion attention to the ordinal scale of the class. Our model with OSLSP outperforms large-scale pre-trained and fine-tuning models in classification tasks of skeletal muscle recovery stages.

\keywords{LLP \and Skeletal muscle \and WSI \and Weakly supervised learning }
\end{abstract}
%
%======================main======================%
%
\renewcommand{\thefootnote}{\arabic{footnote}}
\input{1_introduction}

\input{2_relatedwork}
\input{3_method}
\input{4_result}
\input{5_discussion}
\input{6_Acknowledgements}
\bibliographystyle{splncs04}
\bibliography{9_reference}

\end{document}

%% file: 99_MatsushitaLabTemplate.tex
\usepackage{xspace}
\makeatletter
\DeclareRobustCommand\onedot{\futurelet\@let@token\@onedot}
\def\@onedot{\ifx\@let@token.\else.\null\fi\xspace}

\makeatother

\DeclareMathAlphabet{\mathcal}{OMS}{cmsy}{m}{n}

\newcommand{\fref}[1]{Fig.~\ref{#1}}

\usepackage[]{xcolor}
\newcounter{todos}
\AtEndDocument{\ifnum\value{todos}>0 \PackageWarning{TODOS}{There are \arabic{todos} todos left in this paper! Fix them before submitting the paper!} \fi}

\newcommand{\V}[1]{\ensuremath{\mathbf{#1}}}

\newcommand{\vp}{\V{p}}

\newcommand{\vx}{\V{x}}
\newcommand{\vy}{\V{y}}

%% file: 1_introduction.tex
%LLP in muscle tissue
%Update of feature extractor in %LLP with class order and sim prop

\section{Introduction}
%background and goal
Evaluating the regeneration process of damaged muscle tissues is a fundamental analysis for measuring the effect size of biological experimental manipulations for discovering mechanisms of muscle weakness associated with aging and disease. One method to induce damage and recovery of muscle tissues is an injection of cardiotoxin (CTX), where CTX is injected into mice's lower leg muscles to cause local necrosis of myofibers within the tissue. In recovery, damaged-myofiber-derived factors activate muscle satellite cells, which proliferate and become myoblasts. Many myoblasts fuse to form myotubes, and as they mature, they become embedded in the tissue and return to myofibers.
The recovery speed of cells and fibers in muscle tissue exhibits locality~\cite{Fukada_adiponectin}, requiring the evaluation of the condition of cells in each specific region to assess tissue recovery~\cite{CSA-4-openCSAM}. This process is labor-intensive because muscle tissue images are high-resolution whole slide images (WSI) containing numerous cells and fibers. Therefore, effective and objective automated image analysis is necessitated.

%2 problem nominal and cannot update backbone
Although supervised learning, as represented by deep learning, is a useful method, annotating recovery stages in every region of WSI is similarly labor-intensive and requires specialized expertise. In this context, the classification of multiple instances by learning label proportion (LLP) is one of the leading methods of weakly supervised learning in WSI analysis because it avoids GPU memory limitations caused by the large size of WSI and enhances the explainability of image analysis by predicting each region of the WSI. Examples of applying LLP to WSI include binary~\cite{LLP-IEEE} or three-class~\cite{Matsuo2024LearningFP} necrosis determination in tumors and screening for regions of interest~\cite{LLP-Screening_for_ROIs}. LLP currently faces two challenges in evaluating the cell regeneration process. (1) As LLP computes the loss function after predicting all instances in a bag, it typically uses pre-trained models as feature extractors (backbones) and only updates the classification layer (head)~\cite{LLP-IEEE,Matsuo2024LearningFP}. Large pre-trained models such as DINO~\cite{DINO-origin} are generally trained on common objects, so they may not perform well in extracting features for medical WSI tasks. (2) Cells in the muscle recovery process gradually change their morphology over time, creating a natural biological order among the stages, which needs to be adequately captured.
%as cells in the muscle recovery process gradually change their morphology over time, there is an inherent biologically derived morphological order among the stages.
Specifically, the process involves the following stages: intact myofiber, ghost fibers (which are the basal membrane remnants post-CTX injection), myotubes (satellite cells that have undergone myogenesis), and recovered myofibers (cells that have fully regenerated). However, existing LLP methods treat the process on a nominal scale, leading to missing ordinal information. Therefore, we aim to address both issues by updating the feature extractor while considering the order of the classes in the training process under the LLP paradigm.
%Therefore, we attempt to tackle both issues by updating the feature extractor with consideration for weighted class order in class proportion learning.

% suggest
In this study, we propose an ordinal scale learning from similarity proportion: OSLSP that uses the date elapsed from CTX injection as weak supervision tied to cell class proportions, viewing morphological changes in cells during recovery as ordinal classes. A similarity proportion loss in OSLSP simultaneously addresses the issues of non-updated feature extractors in LLP and the lack of ordinal information among classes. Similarity proportion loss is computed between two bags, where the ground truth similarity proportion of classes is derived from the binomial theorem based on the class proportions of the two bags. During training, the similarity of feature vectors of instances from different bags is calculated, forming a similar distribution when comparing instances from two bags, and the loss function is computed by comparing this to the ground truth proportion. The feature extractor trained with OSLSP contributed to improved accuracy in cell classification tasks in the skeletal muscle recovery dataset.

%% file: 2_relatedwork.tex
\section{Related work}
The weakly-supervised~\cite{pseudo_WSI,LLP-2008-origin,shao2021transmil,IIB-MIL} methods to address the challenges of WSI are shown in Table.\ref{table:relatedwork-WSI}. Previous studies have designed loss functions using class proportions for bag-level learning and generated pseudo-labels for each instance from class proportions for instance-level learning\cite{IIB-MIL}. While instance-level learning allows for the updating of feature extractors, it has been pointed out that pseudo-labels can lead to noise due to incorrect label generation \cite{IIB-MIL}. Additionally, previous bag-level learning approaches have treated the output results as nominal scales, focusing on binary tumor classification \cite{IIB-MIL,pseudo_WSI,shao2021transmil}, ternary classification \cite{Matsuo2024LearningFP}, and the extraction of regions of interest in pathology images \cite{LLP-Screening_for_ROIs}. They have not targeted morphology changes of cells with ordinal scale classes in muscle tissues. Although not targeting WSI, Jerónimo's \cite{LLP-3:embryo} used already quantified patient data and implantation rates for embryo implantation prediction. LLP-VAT \cite{LLP-VAT} proposes a consistency loss to ensure class prediction consistency even with slight noise added to individual instance images, in addition to the traditional LLP loss. SIM-LLP \cite{LLP-SIM} added a pairwise similarity-based loss that penalizes different predictions of feature vectors in addition to the conventional LLP loss. Previous LLP methods have used pre-trained encoders for feature extraction, focusing on learning the classification head. This study aims to update the feature extraction backbone using similarity proportion loss.

%MIL
\begin{table}
    \centering
    \caption{This table lists methods for addressing WSI classification tasks. \emph{Class Scale} indicates the output scale by a classifier head. \emph{Learning Target} denotes the layers of the model updated, where \emph{Backbone} refers to the layer responsible for feature extraction, and \emph{Head} refers to the layer that estimates the class from the extracted features. \emph{Loss} indicates the loss function used for learning. \emph{Prop.} stands for proportion.
    }
    \begin{tabular}{c|c|c|c}
        \toprule
        Method & Class Scale & Learning Target & Loss\\
        \midrule
        Pseudo~\cite{pseudo_WSI} & Nominal & Head & Pseudo Class Label + Class Prop.\\
        LLP~\cite{LLP-2008-origin} & Nominal & Head & Class Prop.\\
        MIL~\cite{shao2021transmil} & Binary & Head & Class Prop.\\
        IIB-MIL~\cite{IIB-MIL} & Binary & \textbf{Backbone} + Head & Pseudo Class Label + Class Prop.\\
        \textbf{OSLSP (Ours)} & \textbf{Ordinal} & \textbf{Backbone} + Head & \textbf{Similarity Prop.} + Class Prop.\\
        \bottomrule
    \end{tabular}
    \label{table:relatedwork-WSI}
\end{table}

%% file: 3_method.tex
\section{Method} 
%======================oslsp======================%
\subsection{OSLSP: Ordinal scale learning from similarity proportion}
\label{sec.sim_prop}
%method killer figure
\begin{figure}[t]
    \includegraphics[width=\linewidth]{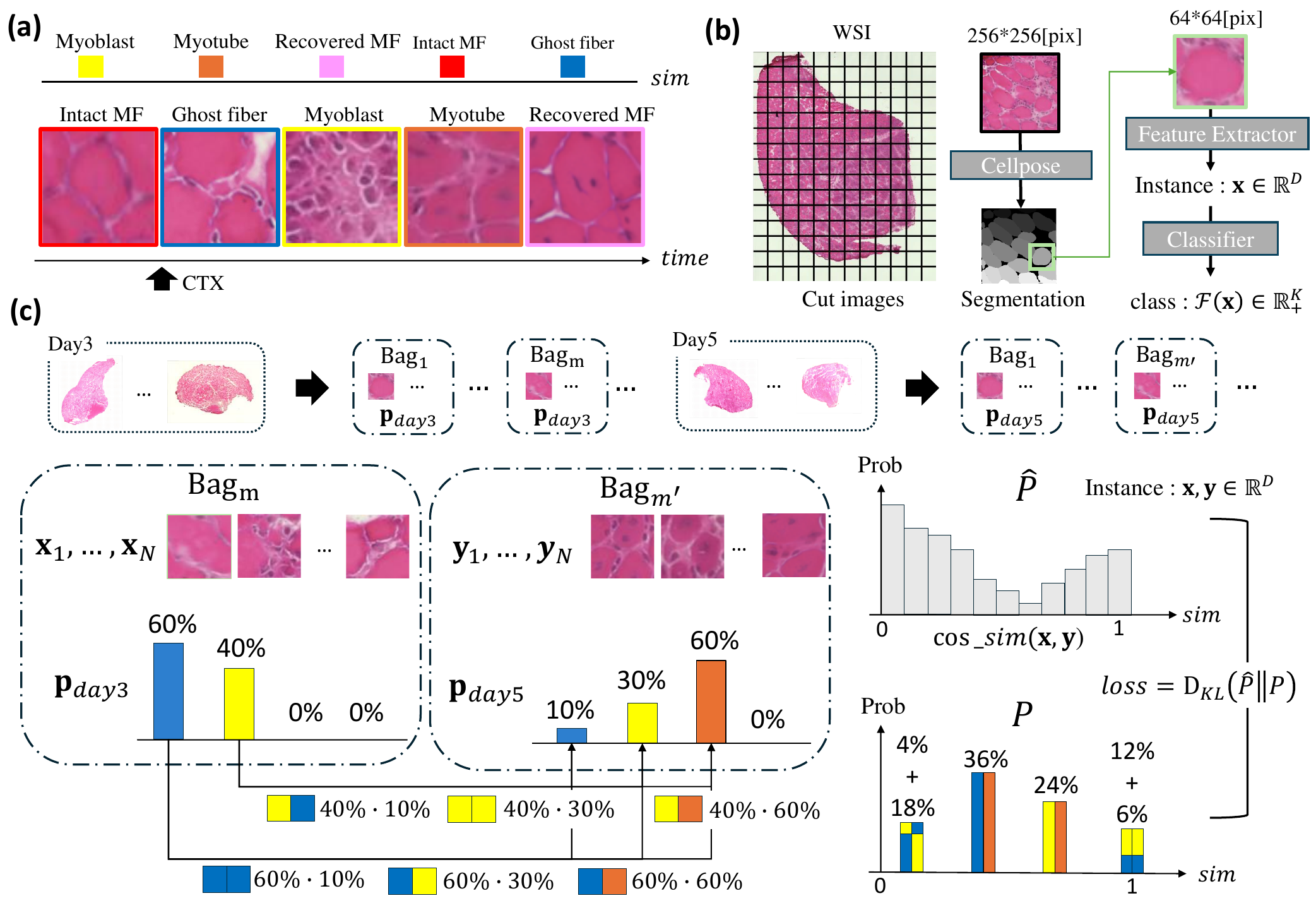}
    \caption{Overview of our OSLSP. \textbf{(a)} Similarity of class and morphological changes of cells over time. \textbf{(b)} A pipeline to obtain instance features and class inference results for each cell from WSI. \textbf{(c)} We computed similarity proportion loss using KL divergence between ground truth similarity distribution derived by combining two bags' proportions and predicted one derived by cosine similarity of each instance in two bags.
    }
    \label{fig:method_osllp}
\end{figure}

We propose a similarity proportion loss to simultaneously address the issues of updating the feature extractor and missing ordinal class information in LLP.
As shown in Fig.~\ref{fig:method_osllp}(a), we define classes $C = {1, ..., k, ... K}$  for different cell morphologies: Red: Intact Myofiber (Intact MF), Blue: Ghost fiber, Yellow: Myoblast, Orange: Myotube, Pink: Recovered Myofiber (Recovered MF). The dates after CTX injection are denoted as $d \in \{``day0", ``day3", ``day5", ``day7", ``day14"\}$, where Day 0 represents the state before CTX injection, meaning 100\% cells are Intact MF. Ghost Fiber is the cell stage where myofibers transition immediately after CTX injection, with the highest proportion on Day 3. Over time, the proportion shifts towards myoblasts and myotubes as days pass. The necrotic cells lose all morphology of the ex-myofiber as shown in ~\fref{fig:method_osllp}(a) ``Myoblast''. Therefore, we see myoblasts as the birth of myofibers, while ghost fibers represent the end of them, making their similarity the most distant, as illustrated in~\fref{fig:method_osllp}(a) ``sim''. At the stage where myoblasts have sufficiently proliferated, they become multinucleated myotubes, which means from Day 5 to Day 7, the proportion of myotubes increases. As myotubes grow, they fill the gaps between cells, resulting in Recovered MF. From this prior knowledge and observation of cell morphology, we perform rough manual class annotation on the training data, obtaining proportions $\vp_d \in [0,1]^K$ relative to each date label $d$.

After dividing the WSI into $256 \times 256$ [pixel] patch images, we use Cellpose~\cite{cellpose-origin} for cell segmentation and clip $64 \times 64$ [pixel] images with each segmented cell in the center for inference, as shown in Fig.~\ref{fig:method_osllp}(b). We obtain instance feature vectors $\vx \in \mathbb{R}^D$($D \in \mathbb{N}$: the dimension of the image feature map) from $64 \times 64$ [pixel] images through a feature extractor (backbone). The classifier model produces inference output $\mathcal{F}: \mathbb{R}^D \rightarrow [0,1]^K (\sum_{k=1}^{K} \mathcal{F}(\vx)_k = 1)$, where $\mathcal{F}(\vx)_k$ represents the classifier's confidence in class $k$. 

Following the standard bag-making method~\cite{llp2023_review} of LLP, we group $N$ instances $\vx \in \mathbb{R}^D$ of the same date into one bag, which has a corresponding class proportion $\vp_d \in [0,1]^K$ ($ \|\vp_{d}\|_1=1, \|\|_1$ means L1 norm), $N$ is defined as the bag size. As shown in Fig.~\ref{fig:method_osllp}(c), we computed the similarity proportion loss by calculating Kullback–Leibler divergence between ground truth probability density function (PDF) $P$ and predicted PDF $\hat{P}$. $P$ is calculated by combining the ground truth class proportion of two bags, and $\hat{P}$ is computed as the cosine similarity between instance features in two bags. Given $b$ as the number of bins of the discretized PDFs, the similarity proportion loss is defined as

%KL
\begin{equation}
    \mathcal{L}_{SimProp} =D_{\text{KL}}(\hat{P} \| P)= \sum_{i=1}^{b} \hat{P}(i) \log\left(\frac{\hat{P}(i)}{P(i)}\right)
\end{equation}

We compute the predicted PDF $\hat{P}$ by randomly picking two bags and compare the instance pairs $\vx_n, \vy_n (n \in \{1,2,...N\})$ of the two bags using a scaled cosine similarity $\text{CosSim}(\vx_n, \vy_n)$: $\mathbb{R}^D \times \mathbb{R}^D \rightarrow [0,1] \in \mathbb{R}$, which is defined as below. 

%cos sim equation
\begin{equation}
    \text{CosSim}(\vx_n, \vy_n) = \frac{1}{2}( \frac{\vx_n \cdot \vy_n}{\|\vx_n\| \cdot \|\vy_n\|} + 1) \in [0,1] 
\end{equation}

We obtain the cosine similarities $\text{CosSim}(\vx_n, \vy_n)$ for all instance pairs and plot them as a histogram. The parameters \emph{max}, \emph{min}, and $b$ are used to set the maximum and minimum values of the data and the number of bins in the histogram. Since the cosine similarity $\text{CosSim}$ values range from 0 to 1, we set $max = 1$ and $min = 0$. We compute the width of the histogram bins $\Delta = \frac{max-min}{b}$. The value of the $i$-th bin of the histogram $\text{hist}(i)$ is determined as 

%compute hist
\begin{equation}
\label{eq:hist}
    \text{hist}(i) = \sum_{n=1}^{N} \text{I}(\text{CosSim}(\vx_n, \vy_n) \in [i \times \Delta, (i+1) \times \Delta])
\end{equation}
where $\text{I}$ is an indicator function that counts values within the specified range. When we use the indicator function $\text{I}$ in the design of the loss function, it results in a non-differentiable computation process, preventing the backpropagation of gradients from the loss to the model. We convert inter-atomic distances from discrete to continuous~\cite{gaussian_feature_expansion} to address this issue. We adopt Gaussian expansion with $\sigma=0.1$ as described below instead of using Eq.~\ref{eq:hist}. 

%Histogram calculation approximation using Gaussian expansion method
\begin{equation}
\label{eq:Gaussian expansion}
    \text{hist}(i) \cong \hat{P}(i) = \sum_{n=1}^{N} \frac{1}{\sigma\sqrt{2\pi}} \exp\left(-\frac{(\text{CosSim}(\vx_n, \vy_n) - \mu_i)^2}{2\sigma^2}\right) \Delta 
\end{equation}
The histogram calculation is approximated in Gaussian expansion by integrating the Gaussian density functions for each data point $i \in \{1,2,...,b\}$ over the bin width. Note that $\mu_i$ represents the mean similarity value of the $i$-th bin. This approach allows for a differentiable histogram calculation, enabling the gradient backpropagation necessary for training the model.

For any two classes $k,k' \in C=\{1,...,K\}$, we define a class similarity as described below for computing ground truth PDF $P$ following the life cycle of myofibers, as shown in Fig.~\ref{fig:method_osllp}(a). When comparing the same class $(k = k')$, their similarity $\text{sim}(k, k')$ equals $1$.

\begin{equation}
\label{eq:sim}
    \text{sim}(k, k')=1 -\frac{|k'-k|}{K-1} \in [0,1]
\end{equation}

Using pair of the class proportions $\vp_{d}, ~\vp'_{d'} \in [0,1]^K$ corresponding to date labels $d$ and $d'$ of the two bags, the ground truth PDF is calculated by

%compute true distribution
\begin{equation}
\label{eq:TrueP}
   \begin{cases}
       P(\text{sim}(k, k')) = p_{k} \cdot p'_{k'} & ~\text{if} \ k = k';
       \\
       P(\text{sim}(k, k')) = p_{k} \cdot p'_{k'} + p_{k'} \cdot p'_{k} & ~\text{otherwise.} 
   \end{cases}
\end{equation}

%======================llp======================%
%\noindent{\bf LLP learning for classifier.}
\subsection{LLP learning for classifier}
To train a classifier head, we compute a proportion loss by calculating KL divergence between ground truth class proportions $\vp_d \in [0,1]^K$ and predicted one $\hat{\vp}_{d} \in [0,1]^K$ as below.

\begin{equation}
    \mathcal{L}_{prop} = D_{\text{KL}}( \vp_d \| \hat{\vp}_{d}) = \sum_{k=1}^{K} p_d \log\left(\frac{p_d}{\hat{p}_d}\right)
\end{equation}

$\mathcal{L}_{prop}$ is computed for each bag, which have $\mathbb{B}=\{\vx_1, \vx_2,...,\vx_N\}$ and class proportion $\vp_d \in [0,1]^K (\|\vp_{d}\|_1=1)$ corresponding date labels $d$. Through the classifier $\mathcal{F}: \mathbb{R}^D \rightarrow \mathbb{R}^K_+$, we obtain the class confidence $\mathcal{F}(\vx) \in [0,1]^K$ and we obtain the predicted class proportions $\hat{\vp}_{d}=[\hat{p}_1,...,\hat{p}_k,..., \hat{p}_K] \in [0,1]^K, \|\hat{\vp}_{d}\|_1=1$ by aggregating the class results of each instance at the bag level as below.

\begin{equation}
    \hat{p}_k = \frac{1}{|\mathbb{B}|} \sum_{\vx \in \mathbb{B}} \mathcal{F}(\vx)_k
\end{equation}

%% file: 4_result.tex
\section{Evaluation}
\noindent{\bf Experiment settings.}
In our evaluation, we use $31$ WSIs and obtain ground truth by professional annotation into $5$ classes, one WSI per day. Thus, we have $26$ WSIs for training and $5$ WSIs for testing. 
Please refer to the supplementary materials for more details.
We augment the data by rotation, flipping, and simulation of random optical conditions, including RandomBrightness (p=0.5), RandomContrast (p=0.5), and RandomGamma (p=0.5) on albumentations v1.3.1 after cutting to $256 \times 256$ [pixel]. To correctly reflect the proportion in the dataset, we randomly clip $64 \times 64$ images from the $256 \times 256$ pixels images for training data. We use ViT-B/8 model\footnote{Self-Supervised Vision Transformers with DINO.~<\url{https://github.com/facebookresearch/dino}>, last accessed on June 29, 2024.}
, a vision transformer model with a patch size of 8 as the backbone. We use the student checkpoint and enable the average pooling patch tokens. Other parameters are kept in default. 
In OSLSP, we only fine-tune the last block of the model. 
In DINO fine-tuning, we follow the official training pipeline and start the training from the pre-trained ViT-B/8 model.
In training a classifier head by LLP using the class proportions of dates after preparing the feature extractor, we use a three-layer perceptron~\cite{rosenblatt1958perceptron} with ReLU~\cite{ReLU} activations.
During inference, $64 \times 64$ images were clipped from the test data centered around cells based on segmentation obtained using a Cellpose~\cite{cellpose-origin} model, fine-tuned with training data on top of the Cyto model. 
We compare our OSLSP with a pre-trained DINO model~\cite{DINO-origin} and a DINO model~\cite{DINO-origin} fine-tuned with our training data. 

\begin{figure}[t!]
    \includegraphics[width=\linewidth]{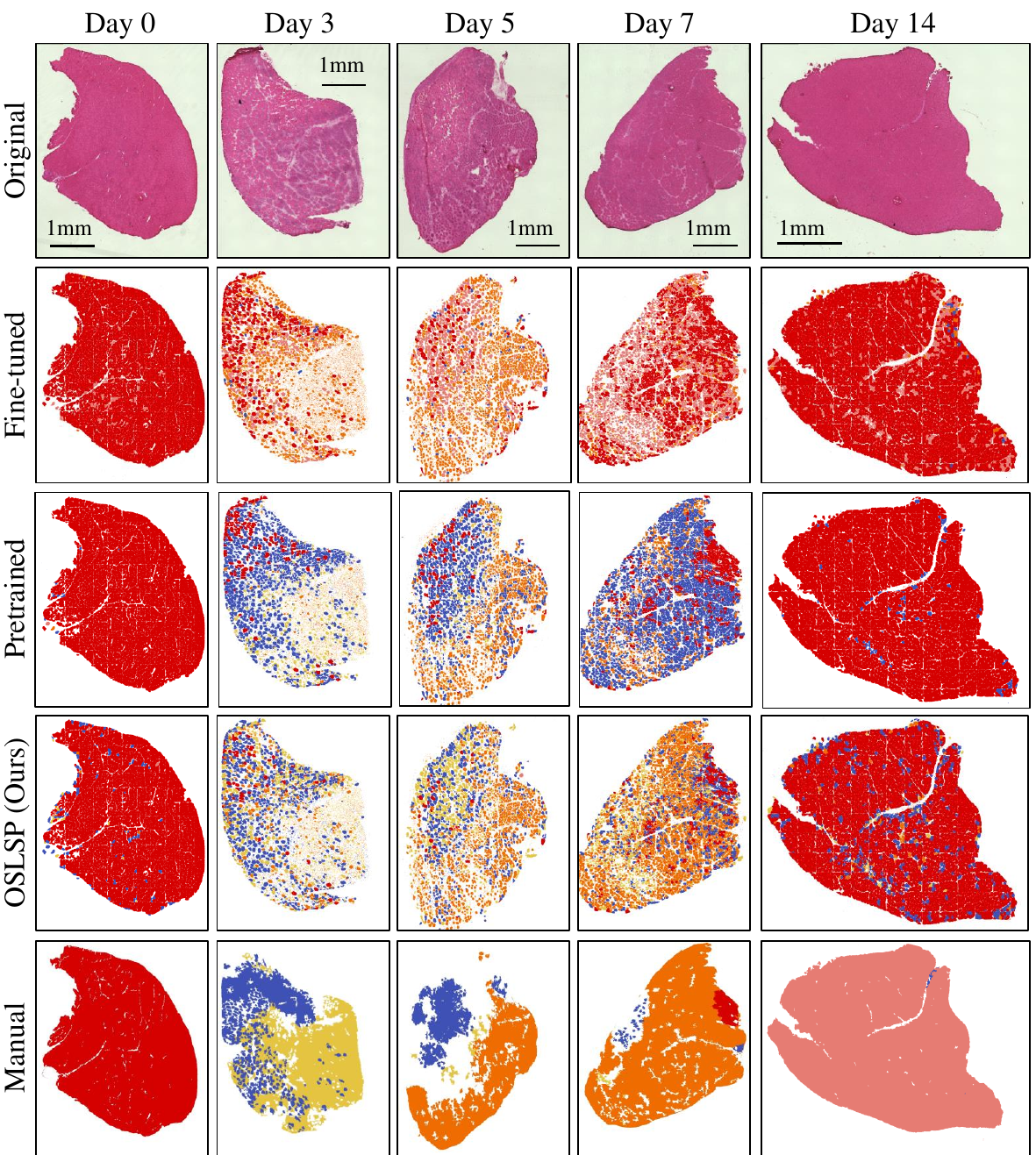}
    \caption{Classification results of WSIs for each day. Blue: ghost fiber, Red: intact myofiber, Pink: recovered myofiber, Orange: myotube, Yellow: myoblast. In the manual expert annotations, unannotated white areas indicate uncertain regions.}
    \label{fig:result_coloring}
\end{figure}

\noindent{\bf Classification results.}
As shown in Table~\ref{table:classifier_result}, we compare OSLSP with the baselines using manual expert annotations as ground truth. RMSE reflects the weight of incorrect predictions in class order ``blue, red, pink, orange, yellow'' as in Fig.~\ref{fig:method_osllp}(a). Fig.~\ref{fig:result_coloring} shows the visual result of OSLSP and our baselines. The empty spaces of Manual in Fig.~\ref{fig:result_coloring} indicate areas where even experts could not indeterminate cell class. When performing the quantitative evaluation of Table ~\ref{table:classifier_result}, the cells in this area are excluded from the calculations.

%method killer figure

\begin{table}[b!]
    \centering
    \caption{Classification results of LLP classifiers. Micro accuracy, macro precision, recall, and F1-score. RMSE: Root mean square error based on ordinal cell class.}
    \begin{tabular}{c|c|c|c|c|c}
        \toprule
        Method &  Accuracy [\%] $\uparrow$ 
        & \hspace{1pt} Recall $\uparrow$\hspace{1pt}  
        & \hspace{1pt} Precision $\uparrow$ \hspace{1pt}
        & \hspace{1pt} F1-score $\uparrow$ \hspace{1pt} 
        & \hspace{1pt} RMSE $\downarrow$ \\
        \midrule
        DINO Pre-trained~\cite{DINO-origin} & $44.442$ & $ 0.436$ & $0.332$ & $0.377$ & $2.431$\\
        DINO Fine-tuned~\cite{DINO-origin} & $20.967$ & $0.251$ & $0.191$ & $0.217$ & $\mathbf{1.76}$\\
        OSLSP (Ours)& $\mathbf{46.005}$ & $\mathbf{0.492}$ & $\mathbf{0.375}$ & $\mathbf{0.425}$ & $2.152$ \\
        \bottomrule
    \end{tabular}
    \label{table:classifier_result}
\end{table}

%We obtain instance $\vx \in \mathbb{R}^D$ by using a feature extractor, which includes a pre-train model of DINO\cite{DINO-origin}, fine-tuning model by DINO method, and OSLLP model for k-Nearest Neighbors.

%% file: 5_discussion.tex
\section{Discussions and Conclusion}
%The interesting aspects of our method are twofold.
\noindent{\bf Discussions.} Applying the concept of contrastive learning to LLP enables the updating of feature extractors using class proportion as weak supervision. OSLSP reduces the domain gap of images between training and test data more effectively than the simple fine-tuned DINO model and better captures skeletal muscle cell features. 
In the supplementary, we include the UMAP plots for all feature extractors for better illustrations. 
As a result, we observe improvements in both RMSE and accuracy for OSLSP in Table~\ref{fig:result_coloring}, which indicates that OSLSP learns the order of the classes more correctly. In contrast, although the fine-tuned model exhibits a low RMSE, it converges to predicting mostly intact MF, recovered MF, and myotubes (as in~\fref{fig:result_coloring}), significantly harming the accuracy.

Unlike traditional LLP that treats classes as nominal scales, OSLSP considers the ordinal scale nature of class similarity, given that intact MF and ghost fibers are in similar states immediately before and after CTX injection. Our approach is similar to applying class weighting in supervised learning~\cite{ordinal_distribution_2017deep}, which assigns different levels of importance to errors in the estimation during training. One of our limitations, as shown in Fig.~\ref{fig:result_coloring}, is that intact MF and ghost fiber are more frequently confused than the pre-trained model due to the high pre-set similarity of the two classes. Care must be taken with the weighting between these classes, as intact MF and necrotic ghost fibers, while morphologically similar, are totally different regarding muscle strength. Therefore, misclassifying the two classes might have a substantial negative impact when measuring muscle strength. 

\noindent{\bf Conclusion.}
In this study, we propose that OSLSP update a feature extractor by a similarity proportion loss, allowing the injection of prior knowledge into class order. OSLSP enables ordinal scale learning tailored to the objective and achieves better analysis of muscle recovery. 

%% file: 6_Acknowledgements.tex
\section{Acknowledgements}
JSPS KAKENHI Grant Numbers JP22H05085 and JP22K12246 partially supported this work. 